\documentclass[10pt,a5paper,twoside]{article}
\usepackage{coling2012}
\usepackage{amsmath}
\usepackage{multirow} 
\usepackage{rotating}
\usepackage{array}
\title{Authorship Identification in Bengali Literature: a Comparative Analysis }

\author{$Tanmoy\ Chakraborty$\\
{\small Department of Computer Science \& Engineering\\
Indian Institute of Technology, Kharagpur\\
India
 		\\
  \texttt{its\_tanmoy@cse.iitkgp.ernet.in} \\ 
}}

\begin{document}
\maketitle
  
\abstractEn{ 
\\
Stylometry is the study of the unique linguistic styles and writing behaviors of individuals. It belongs to the core task of text
categorization like authorship identification, plagiarism detection etc. Though reasonable
number of studies have been conducted in English language, no major work has been done so far in Bengali. In this work, We will present a
demonstration of authorship identification of the documents written in Bengali. We adopt a set of fine-grained stylistic
features for the analysis of the text and use them to develop two different models:
statistical similarity model consisting of three measures and their combination, and machine learning model with Decision Tree, Neural
Network and SVM. Experimental results show that SVM outperforms other state-of-the-art methods after 10-fold cross validations. We also
validate the relative importance of each stylistic feature to show that some of them remain consistently significant in every model used in
this experiment.
}

\keywordsEn{Stylometry, Authorship Identification, Vocabulary Richness, Machine Learning}

\newpage

\section{Introduction}

Stylometry is an approach that analyses text in text mining e.g., novels, stories, dramas that
the famous author wrote, trying to measure the author’s style, rhythm of his pen, subjection of his desire, prosody of his mind by choosing
some attributes which are consistent throughout his writing, which plays the linguistic fingerprint of that author. Authorship
identification belongs to the subtask of Stylometry detection where a
correspondence between the predefined writers and the unknown articles has to be established taking into account various stylistic features
of the documents. The main target in this study is to build a decision making system that enables users to predict and to choose the right
author from a specific anonymous authors' articles under consideration, by choosing various lexical, syntactic, analytical features called
as \textit
{stylistic markers}. Wu incorporate two models\textemdash (i) statistical model using three well-established similarity measures-
cosine-similarity, chi-square measure, euclidean
distance, and (ii) machine learning approach with Decision Tree, Neural Network and Support Vector Machine (SVM). 

The pioneering study on authorship attributes identification using word-length histograms appeared at the very end of nineteen
century~\cite{Malyutov}. After that, a number of studies based on content analysis~\cite{krippendorff}, computational stylistic
approach~\cite{stamatatos}, exponential gradient learn algorithm ~\cite{Argamon2003Style}, Winnow regularized algorithm~\cite{ZhangDJ02},
SVM based approach~\cite{Pavelec} have been proposed for various languages like English, Portuguese (see~\cite{Stama} for reviews). As a
beginning of Indian language Stylometry analysis, ~\cite{Chanda} started working with handwritten Bengali texts to judge authors. 
~\cite{Das} proposed an authorship identification task in Bengali using simple n-gram token counts. Their approach is restrictive when
considering authors of the same period and same genre. The texts we have chosen are of the same genre and of the same
time period to ensure that the success of the learners would infer that texts can be classified only on the style, not by the prolific
discrimination of text genres or distinct time of writings. We have compared our methods with
the conventional technique called \textit{vocabulary richness} and the existing method proposed by ~\cite{Das} in Bengali. The observation
of the
effect of each stylistic feature over 10-cross validations relies on that fact that some of them are inevitable for authorship
identification task especially in Bengali, and few of the rare studied features could accelerate the performance of this mapping task.   
 


\section{Proposed Methodology} 
\begin{figure}[h!]

\centering
\setlength\fboxsep{7pt}
\setlength\fboxrule{0.5pt}
\fbox{\includegraphics[width=70 mm]{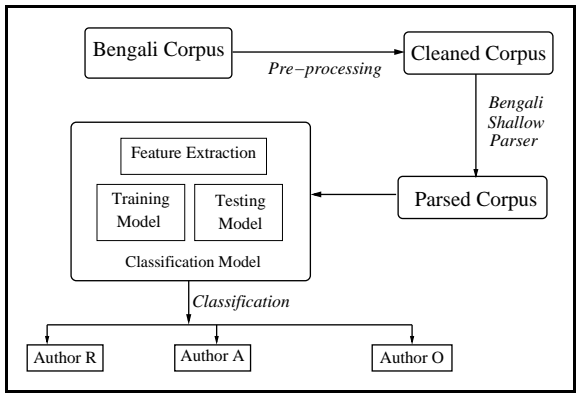}}
\caption{System architecture}
\label{model}
\end{figure}

The system architecture of the proposed
stylometry detection system is shown in Figure~\ref{model}. In this section, we briefly describe different
components of
the system architecture and then analytically present the set of stylistic features.

\subsection{Textual analysis} Basic pre-processing before actual textual analysis is required so that stylistic markers are clearly viewed
to the system for further analysis. Token-level markers discussed in the next subsection are extracted from this pre-processed corpus.
Bengali Shallow parser\footnote{http://ltrc.iiit.ac.in/analyzer/bengali} has been used to separate the sentence and the chunk boundaries and
to identify parts-of-speech of each token. From this parsed text, chunk-level and context-level markers are also demarcated.

\subsection{Stylistic features extraction}
Stylistic features have been proposed as more reliable style markers than for example, word-level features since the stylistic markers are
sometime not under
the conscious control of the author. To allow the selection of the linguistic features rather than n-gram terms, robust and accurate text
analysis tools such as lemmatizers, part-of-speech (POS) taggers, chunkers etc are needed. We have used the Shallow parser, which gives a
parsed output of a raw input corpus. The stylistic markers which have been selected in this experiment are discussed in
Table~\ref{table:feature}. Most of the features described in Table~\ref{table:feature} are self-explanatory. However, the problem occurs
when identifying keywords (KW)
from the articles of each author which serve as the representative of that author. For this, we have identified top fifty high frequent
words (since we have tried
to generate maximum distinct and non-overlapped set of keywords) excluding stop-words in Bengali for each author using $TF\ast
IDF$ method. Note that, all the features are normalized to make the system independent of document length.

\begin{table}[!h]
\setcounter{table}{0}
\small{
 \centering{
\begin{tabular}{|c|c|c|c|c|}
 \hline
 & \textbf{No.} & \textbf{Feature}& \textbf{Explanation}& \textbf{Normalization}\\\hline
\multirow{13}{5 mm}{\begin{sideways}{\textbf{Token Level}}\end{sideways}}
 & 1.& L(w) & Average length of the word & Avg. len.(word)/ Max len.(word)\\\cline{2-5}
 & &  & Intersection of the keywords   & \\
 & 2.  & $KW(R)$    & of Author R and the test &  $|KW(doc)\bigcap KW(R)|$                                \\
 &   &     & document                   &                                  \\\cline{2-5}
 & &  & Intersection of the keywords   & \\
 & 3.  & $KW(A)$    & of Author A and the test &  $|KW(doc)\bigcap KW(A)|$                                \\
 &   &     & document                   &                                  \\\cline{2-5}
 & &  & Intersection of the keywords   & \\
 & 4.  & $KW(O)$    & of Author O and the test & $|KW(doc)\bigcap KW(O)|$                                 \\
 &   &     & document                   &                                  \\\cline{2-5}
 & 5.& HL& Hapex Legomena (No of           & count(HL)/count(word) \\
 &   & & words with frequency=1)             &     \\\cline{2-5}
 & 6.& Punc. & No of punctuations & count(punc)/count(word)\\\hline

\multirow{5}{5 mm}{\begin{sideways}{\textbf{Phrase Level}}\end{sideways}}
 & 7. & NP & Detected Noun Phrase & count(NP)/count of all phrase\\\cline{2-5}
 & 8. & VP & Detected Verb Phrase & count(VP)/count of all phrase\\\cline{2-5}
 & 9. & CP & Detected Conjunct Phrase & count(CP)/count of all phrase\\\cline{2-5}
& 10. & UN & Detected unknown word & count(POS)/count of all phrase\\\cline{2-5}
& 11. & RE & Detected reduplications & count(RDP+ECHO)/count of\\
&     &        & and echo words         & all phrase\\\hline

\multirow{6}{5 mm}{\begin{sideways}{\textbf{Context Level}}\end{sideways}}
& 12.& Dig & Number of the dialogs & Count(dialog)/ No. of\\
&    &       &                           & sentences\\\cline{2-5}
& 13.& L(d) & Average length of the dialog & Avg. words per dialog/ No. of\\
&    &       &                           & sentences\\\cline{2-5}
& 14.& L(p) & Average length of the & Avg. words per para/ No. of\\
&    &       &          paragraph       & sentences\\\hline

\end{tabular}
}
\caption{Selected features used in the classification model}\label{table:feature}

}
\end{table}

\subsection{Building classification model}
Three well-known statistical similarity based metrics namely Cosine-Similarity (COS), Chi-Square measure (CS) and Euclidean Distance (ED)
are used to get their individual effect on classifying documents, and their combined effort (COM) has also been reported. For
machine-learning model, we incorporate three different modules: Decision Trees (DT)\footnote{See5 package by Quinlan,
http://www.rulequest.com/see5-info.html}, Neural Networks
(NN)\footnote{Neuroshell – the commercial software package, http://www.neuroshell.com/} and Support Vector Machine (SVM). For
training and classification phases of SVM, we have used YamCha\footnote{http://chasen-org/ taku/software/yamcha/} toolkit and TinySVM-
0.07\footnote{http://cl.aist-nara.ac.jp/taku-ku/software/TinySVM} classifier respectively with pairwise multi-class decision method
and the polynomial kernel.

\section{Experimental Results}
\subsection{Corpus}
Resource acquisition is one of the challenging obstacles to work with electronically resource constrained languages like Bengali. However,
this system has used 150 stories in Bengali written by the noted Indian Nobel laureate Rabindranath
Tagore\footnote{http://www.rabindra-rachanabali.nltr.org}. We choose this domain for two reasons: firstly, in such writings the
idiosyncratic style of the author is not likely to be overshadowed by the characteristics of the corresponding text-genre; secondly, in the
previous research~\cite{Tanmoy}, the author has worked on the corpus of Rabindranath Tagore  to explore some of the stylistic
behaviors of his documents. To differentiate them from other authors’ articles, we have selected 150 articles of Sarat Chandra Chottopadhyay
and 150 articles\footnote{http://banglalibrary.evergreenbangla.com/} of a group of other authors (excluding previous
two authors) of the same time period. We divide 100
documents in each cluster for training and validation purpose and rest for testing. The statistics
of the entire dateset is tabulated in Table~\ref{table:corpus}. Statistical
similarity based measures use all 100 documents for making representatives the clusters. In machine learning models, we use 10-fold
cross validation method discussed later for better constructing the validation and testing submodules. This demonstration focuses on two
topics: (a) the effort of many authors on feature selection and learning and (b) the effort of limited data in authorship detection. 

\begin{table}[!h]
 \centering
\small{
\begin{tabular}{|c|c|c|c|c|}
\hline
 Clusters & Authors & No. of documents & No. of tokens & No. of unique tokens\\\hline
 & Rabindranath &  &  & \\
Cluster 1          & Tagore       & 150    & 6,862,580        & 4,978,672       \\
          & (Author R)   &      &        &   \\\hline
 & Sarat Chandra &  &  & \\
 Cluster 2         & Chottopadyhay      & 150    & 4,083,417 & 2,987,450\\        
          & (Author A)  & & &\\\hline
Cluster 3 & Others & 150 & 3,818,216 & 2,657,813\\
	  & (Author O) & & &\\\hline
\end{tabular}
\caption{Statistics of the used dataset}\label{table:corpus}}
\end{table}

\subsection{Baseline system (BL)}
 In order to set up a baseline system, we use traditional
lexical-based methodology called \textit{vocabulary richness} (VR)~\cite{David} which is basically the type-token ratio
$(V/N)$, where $V$ is the size of the vocabulary of the sample
text and $N$ is the number of tokens which forms the simple text.  By using nearest-neighbor algorithm, the baseline system tries to map
each of the testing documents to one author. We have also compared our approach with the state-of-the-art method proposed by~\cite{Das}. The
results of the baseline systems are depicted using confusion matrices in Table~\ref{table:baseline}. 

\begin{table}[!h]
 \centering
\small{
\begin{tabular}{|c|c|c|c|c|c|c|c|c|}
\hline
\multicolumn{5}{|c|}{Vocabulary richness (VR)} &\multicolumn{4}{|c|}{~\cite{Das}}\\\hline
&R & A & O & e(error) in \% & R & A & O & e(error) in \% \\\hline
R& \textit{26} & 14 & 10 & 48\% & \textit{31} & 9 & 10 & 38\%\\\hline
A& 17 & \textit{21} & 12 & 58\% & 18 & \textit{30} & 2 & 40\% \\\hline
O& 16 & 20 & \textit{14}& 72\% & 10 & 6 & \textit{34}& 32\%\\\hline
\multicolumn{4}{|c|}{Avg. error} & 56\% & \multicolumn{3}{|c|}{Avg. error} & 36.67\%\\\hline
\end{tabular}
\caption{Confusion matrices of two baseline system (correct mappings are italicized diagonally).}\label{table:baseline}}
\end{table}

\subsection{Performances of two different models}

The confusion matrices in Table~\ref{stst} describe the accuracy of the statistical measures and the results of their combined voting. The
accuracy of the majority voting technique is 67.3\% which is relatively better than others. Since the
attributes tested are continuous, all
the decision trees are constructed using the fuzzy threshold parameter, so that the knife-edge behavior for decision trees is softened by
constructing an interval close to the threshold. For neural network, many structures of the multilayer network were experimented with before
we came up with our best network. Backpropogation feed forward networks yield the best result with the following architecture: 14 input
nodes, 8 nodes on the first hidden layer, 6 nodes on the second hidden layer, and  6 output nodes (to act as error correcting codes). Two
output nodes are allotted to a single author (this increases the Hamming distance between the classifications - the bit string that is
output with each bit corresponding to one author in the classification- of any two authors, thus decreasing the possibility of
misclassification). Out of 100 training samples, 30\% are used in the validation set which determines whether over-fitting has occurred and
when to stop training. It is worth noting that the reported results are the average of 10-fold cross validations. We will discuss the
comparative results of individual cross validation phase in the next section. Table~\ref{ml} reports the error rate of individual model in
three confusion matrices. At a glance, machine learning approaches especially SVM (83.3\% accuracy) perform tremendously well compared to
the other models.

\begin{table}[!h]
\centering
\small
\tabcolsep 5 pt 
{
\begin{tabular}{|c|c|c|c|c|c|c|c|c|c|c|c|c|c|c|c|c|}
\hline
\multicolumn{17}{|c|}{Statistical similarity models}\\\hline
&\multicolumn{4}{|c|}{Cosine similarity}&\multicolumn{4}{|c|}{Chi-square measure}&\multicolumn{4}{|c|}{Euclidean
distance}&\multicolumn{4}{|c|}{Majority voting}\\
&\multicolumn{4}{|c|}{(COS)}&\multicolumn{4}{|c|}{(CS)}&\multicolumn{4}{|c|}{(ED)}&\multicolumn{4}{|c|}{(COM)}\\\hline
& R & A & O &e(\%)& R & A & O &e(\%)& R & A & O &e(\%)& R & A & O &e(\%)\\\hline
R & \textit{30} & 12 & 8 & 40 & \textit{34} & 9 & 7 & 32 & \textit{27} & 15 & 8 & 46& \textit{34} & 7 &9& 28\\\hline
A & 15 & \textit{27} & 8 & 46 & 14 & \textit{30} & 6 & 40& 18 & \textit{26}& 6 & 48 & 11 & \textit{32} & 7 & 36\\\hline
O & 12 & 9 & \textit{29} & 42 & 9 & 8 & \textit{33} & 34 & 17& 6 & \textit{27}& 46 & 6 & 11 & \textit{33} & 34\\\hline
&\multicolumn{3}{|c|}{Avg. error}&42.7&\multicolumn{3}{|c|}{Avg. error}&35.3&\multicolumn{3}{|c|}{Avg.
error}&46.6&\multicolumn{3}{|c|}{Avg. error}&32.7\\\hline 

\end{tabular}
}
\caption{Confusion matrices of statistical similarity measures on test set.}\label{stst}
\end{table}

\begin{table}[!h]
\centering
\small{
\begin{tabular}{|c|c|c|c|c|c|c|c|c|c|c|c|c|}
\hline

\multicolumn{13}{|c|}{Machine Learning models}\\\hline
&\multicolumn{4}{|c|}{Decision Tree}&\multicolumn{4}{|c|}{Neural Networks}&\multicolumn{4}{|c|}{Support
Vector Machine}\\\hline
& R & A & O &e(\%)& R & A & O &e(\%)& R & A & O &e(\%)\\\hline
R & \textit{35} & 8 & 6 & 28 & \textit{38} & 9 & 3 & 24 & \textit{44} & 3 & 3 & 12\\\hline
A & 7 & \textit{37} & 6 & 26 & 10 & \textit{35} & 5 & 30& 8 & \textit{40}& 2 & 20 \\\hline
O & 6 & 5 & \textit{39} & 22 & 9 & 5 & \textit{36} & 28 & 2& 7 & \textit{41}& 18 \\\hline
&\multicolumn{3}{|c|}{Avg. error}&25.3&\multicolumn{3}{|c|}{Avg. error}&27.3&\multicolumn{3}{|c|}{Avg.
error}&16.7\\\hline 
\end{tabular}
\caption{Confusion matrices of machine learning models on test set (averaged over 10-fold cross validations).}\label{ml}}
\end{table}

\subsection{Comparative analysis}
The performance of any machine learning tool highly depends on the population and divergence of training samples. Limited dataset can
overshadowed the intrinsic productivity of the tool. Because of the lack of large number of dataset, we divide the training data randomly
into 10 sets and use 10-fold
cross validation technique to prevent overfitting for each machine learning model. The boxplot in Figure~\ref{fig:crossvalid}(a) reports the
performance of each model on 10-fold cross validation phrase with mean accuracy and variance. In three cases, since the notches in the box
plots overlap, we can conclude, with certain confidence, that the true medians do not differ. The outliers are marked separately with the
dotted points. The difference between lower and upper quartiles in SVM is comparatively smaller than the others that shows relative low
variance of accuracies in different iterations.  

We also measure the pairwise agreement in mapping three
types of authors using Cohen's Kappa coefficient~\cite{Cohen}. In Figure~\ref{fig:crossvalid}(b), the high correlation between Decision Tree
and Neural Network models, which is considerably high compared
to the others signifies that the effects of both of these models in author-document mapping task are reasonably identical and less
efficient compared to SVM model.

\begin{figure}[h!]\label{fig:crossvalid}
\centering
\includegraphics[width=100 mm]{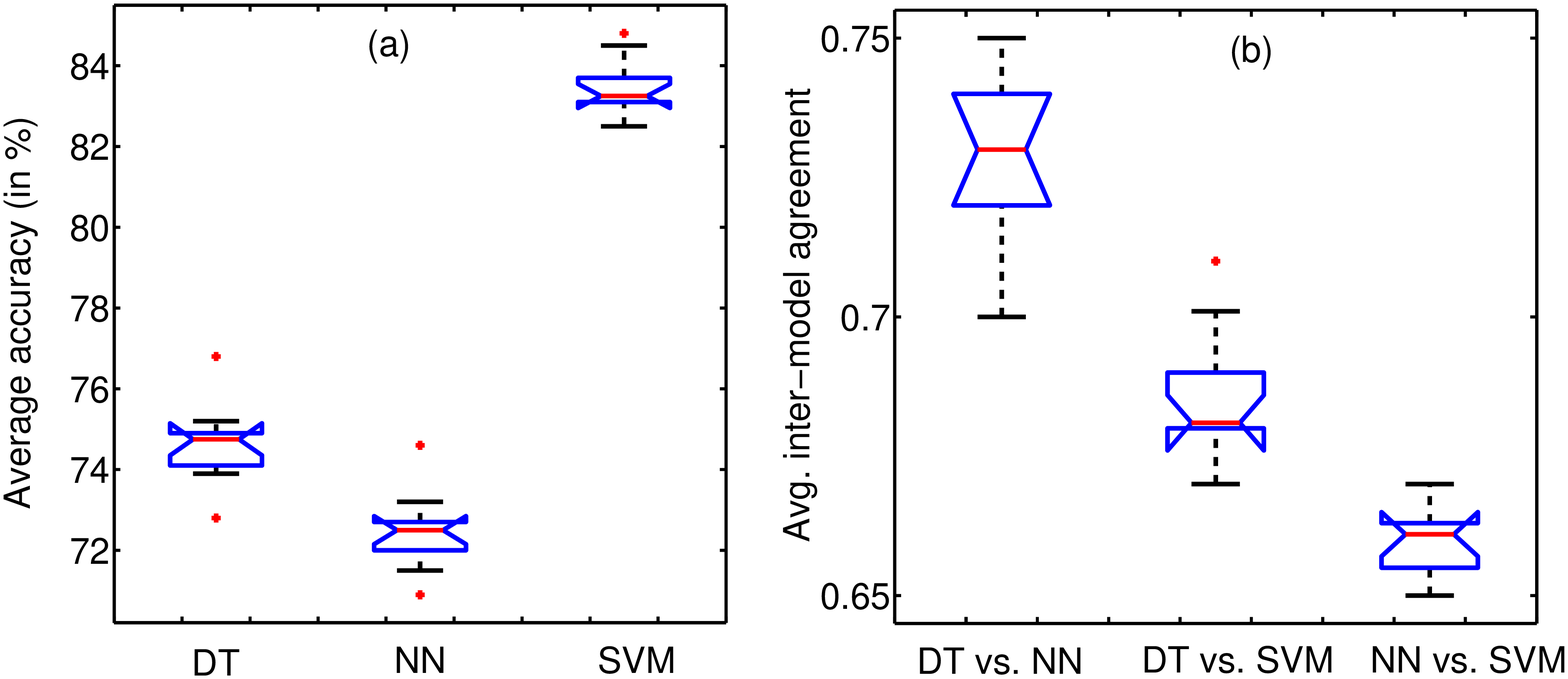}
\caption{(a) Boxplot of average accuracy (in \%) of three machine learning modules on 10-fold cross validations; (b) pair-wise average
inter-model agreement of the models using Cohen's Kappa measure.}
\label{fig:crossvalid}
\end{figure}

\begin{figure}[h!]\label{fig:feature}
\centering
\includegraphics[width=90 mm]{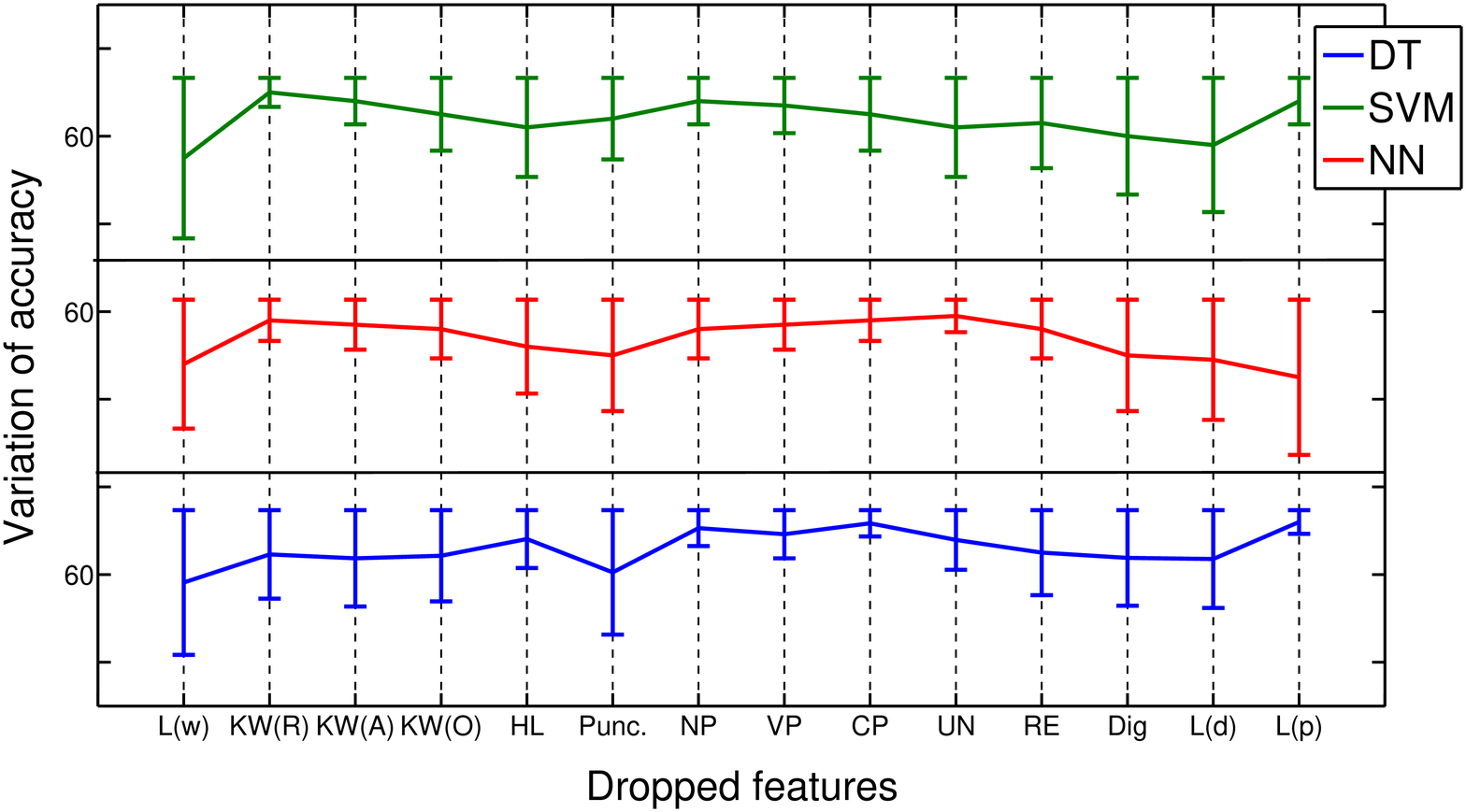}
\caption{(Color online) Average accuracy after deleting features one at a time (the magnitude of the error bar indicates the difference of
the accuracies before and after dropping one feature for each machine learning model).}
\label{fig:feature}
\end{figure}

As a pioneer of studying different machine learning models in Bengali authorship task, it is worth measuring the relative importance
of individual feature in each learning model that gets some features high privilege and helps in feature ranking. We have dropped each
feature one by one and pointed out its relative impact on accuracy over 10-fold cross validations. The points against each feature in the
line
graphs in Figure~\ref{fig:feature} show percentage of accuracy when that feature is dropped, and the magnitude of the corresponding
error bar measures the difference between final accuracy (when all features present) and accuracy after dropping that feature. All models
rely on the high
importance of length of the word in this task. All of them also reach to the common consensus of the importance of KW(R), KW(A), KW(O), NP
and CP.
But few of the features typically reflect unpredictable signatures in different models. For instance, length of the dialog and unknown word
count show larger
significance in SVM, but they are not so significant in other two models. Similar characteristics are also observed in Decision tree and
Neural network models. 

Finally, we study the responsibility of individual authors for producing erroneous results. Figure~\ref{fig:bar} depicts that almost in
every case, the system has little overestimated the authors of documents as author R. It may occur due to the acquisition of documents
because
the
documents in cluster 2 and cluster 3 are not so diverse and well-structured as the documents of Rabindranath Tagore. Developing
appropriate corpus for this study is itself a separate research area specially when dealing with learning modules, and it takes huge amount
of time. The more the focus will be on this language, the more we expect to get diverge corpus of different Bengali writers.

\begin{figure}[h!]\label{fig:bar}
\centering
\includegraphics[width=100 mm]{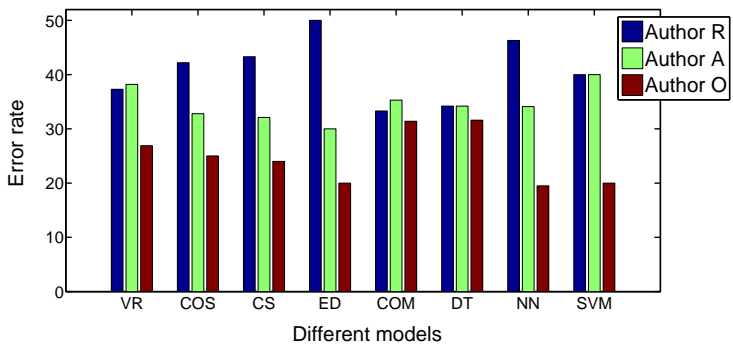}
\caption{(Color online) Error analysis: percentage of error occurs due to wrong identified authors.}
\label{fig:bar}
\end{figure}

\section{Conclusion and Future work}
This paper attempts to demonstrate the mechanism to recognize three authors in Bengali literature based on their style of writing (without
taking into account the author's profile, genre or writing time). We
have incorporated both statistical similarity based measures and three machine learning models over same feature sets and compared them
with the baseline system. All of the machine learning models especially SVM yield a significantly higher accuracy than other models.
Although the SVM yielded
a better numerical performance, and are considered inherently suitable to capture an intangible concept like style, the decision trees are
human readable making it possible to define style. While more features could produce additional discriminatory material, the present study
proves that artificial
intelligence provides stylometry with excellent classifiers that require fewer and relevant input variables than traditional statistics. We
also showed that the significance of the used features in authorship identification task are relative to the used model. This
preliminary study is the journey to reveal the intrinsic style of writing of the Bengali authors based upon which we plan to build more
robust, generic and diverge authorship identification tool. 
\bibliographystyle{apa}

\bibliography{colingbiblio}
\nocite{*}

\end{document}